%% file: main.tex
\documentclass[letterpaper, 10 pt, conference]{ieeeconf}  

\IEEEoverridecommandlockouts                              

\usepackage[dvipdfmx]{graphicx} 
\usepackage{multirow}
\usepackage{url}
\usepackage{spverbatim}
\usepackage{ascmac}
\usepackage{algorithm}
\usepackage{overpic}
\usepackage[noend]{algorithmic}
\usepackage{amsmath}
\usepackage{booktabs}
\usepackage{ascmac}
\usepackage{moreverb}

\usepackage{censor}

\usepackage{tcolorbox}
\tcbuselibrary{breakable} 
\usepackage{mdframed}

\title{\LARGE \bf
From Dialogue to Execution: Mixture-of-Agents Assisted Interactive Planning for Behavior Tree-Based Long-Horizon Robot Execution
}

\author{
Kanata Suzuki$^{1}$,
Kazuki Hori$^{1}$,
Haruka Miyoshi$^{3,4}$,
Shuhei Kurita$^{2,3}$,
and Tetsuya Ogata$^{1,2,3}$
\thanks{
$^{1}$Kanata Suzuki, Kazuki Hori, and Tetsuya Ogata are affiliated with Waseda University, Tokyo 169-8050, Japan. 
$^{2}$Shuhei Kurita and Tetsuya Ogata are affiliated with the National Institute of Informatics (NII), Tokyo 101-8430, Japan. 
$^{3}$Haruka Miyoshi, Shuhei Kurita, and Tetsuya Ogata are affiliated with NII Research and Development Center for Large Language Models. 
$^{4}$Haruka Miyoshi is also affiliated with The Graduate University for Advanced Studies (SOKENDAI), Kanagawa 240-0193, Japan. 
E-mail:{\tt\small suzuki.kanata@aoni.waseda.jp}
}}

\begin{document}

\maketitle
\thispagestyle{empty}
\pagestyle{empty}


\input{sections/0_abstract}

\section{INTRODUCTION}
\label{sec1}

\input{sections/1_introduction.tex}

\section{RELATED WORK}
\label{sec2}

\input{sections/2_related_work.tex}

\section{PROPOSED METHOD}
\label{sec3}

\input{sections/3_method}

\section{EXPERIMENTS}
\label{sec4}
\input{sections/4_experiments}

\section{RESULTS AND DISCUSSION}
\label{sec5}

\input{sections/5_results}

\section{CONCLUSION}
\label{sec6}
\input{sections/6_conclusion.tex}



\section*{ACKNOWLEDGMENT}
This work was supported by JST, CRONOS, Japan Grant Number JPMJCS24K6.
The authors used Claude (Anthropic) for language editing and condensation of the manuscript; it was not used to generate research ideas, experimental data, figures, or results.

\bibliographystyle{IEEEtran}
\bibliography{main}

\end{document}

%% file: sections/0_abstract.tex
\begin{abstract}

Interactive task planning with large language models (LLMs) lets robots generate high-level action plans from natural language, but over long horizons it asks many questions, and tabular plan representations become hard to manage.
We propose a framework that integrates Mixture-of-Agents (MoA)-based proxy answering into interactive planning and generates Behavior Trees (BTs) for structured long-term execution. 
We formulate the MoA as an abstention-based delegation cascade: each expert agent answers only the questions entailed by its own prerequisite description, forwards the rest unchanged, and the human user acts as the terminal fallback.
The question set is thus partitioned disjointly, so no answer fusion or arbitration is required while every question is still resolved.
The BT represents task logic hierarchically and enables retry and dynamic switching among robot policies.
Experiments on a cocktail-making task show that the method removes approximately 27\% of the human responses while keeping the generated BTs within the baseline generator's own variance. 
Real-robot experiments on a smoothie-making task further demonstrate successful long-horizon execution with adaptive policy switching and recovery from action failures. 
We further analyze the failure modes of the framework and show that its applicability boundary is set by the reliability of the weakest action node rather than by the planner.
These results indicate that MoA-assisted interactive planning improves dialogue efficiency while preserving execution quality in real-world robotic tasks.

\end{abstract}

%% file: sections/1_introduction.tex
Long-horizon task execution remains a fundamental challenge in learning-based robotic control: tasks composed of multiple sequential stages with branching and repetition are difficult for models that rely on short-term memory~\cite{act}\cite{dp}.
Large Language Models (LLMs~\cite{chat-gpt}) and Vision-Language Models (VLMs~\cite{Gemini}) have enabled sophisticated language-based planning~\cite{saycan}\cite{codeaspolicies2022}\cite{Vemprala2023}\cite{song2023llmplanner}\cite{yang2024vlmtamp} for such tasks.
Among these, interactive task planning~\cite{Huang2022a}\cite{ren}\cite{hori} is a promising direction: it refines specifications and preconditions through question--answer interaction with the user, producing executable plans from abstract instructions without detailed task engineering in advance.

Two challenges follow.
First, answering every question posed by the LLM planner imposes substantial temporal and cognitive burden, especially over long horizons.
Second, the tabular plan representations commonly used become verbose and hard to manage as complexity grows; language-conditioned imitation learning~\cite{Toyoda2022}\cite{Suzuki2024}\cite{jang2021bc} and Vision-Language-Action (VLA) models~\cite{kim2024openvla}\cite{pi05} likewise accept only a few instructions at a time, and even $\pi_{0.5}$, which folds subtask decomposition into the learning process, remains limited by token length.

For the first, we integrate a Mixture-of-Agents (MoA) into interactive planning to provide proxy responses within the dialogue (Fig.~\ref{fig2}), delegating questions answerable from general knowledge to the agents themselves.
For the second, we adopt Behavior Trees (BTs~\cite{bt-1}\cite{bt-2}), which describe complex action logic hierarchically and modularly and are therefore well suited to long-horizon tasks with repetition and conditional branching.

Our contributions are threefold.
(i) We formulate proxy answering as an \textit{abstention-based delegation cascade}: each agent answers only what its own prerequisites entail and forwards the rest unchanged, so the question set is partitioned disjointly, no answer fusion is required, and the human acts as the terminal fallback that guarantees every question is resolved (Sec.~\ref{sec3}).
(ii) We characterize how this differs from architectures such as layered MoA~\cite{moa}, debate~\cite{debate}, and self-consistency~\cite{selfconsistency}, which aggregate redundant opinions into one answer with no notion of abstention or human fallback (Table I).
(iii) We bind each BT action node to an individual imitation learning policy, and report a real-robot long-horizon task together with an analysis of the failure modes and applicability boundaries of the framework (Sec.~\ref{sec5}).

%% file: sections/2_related_work.tex
LLM-based robot task planning divides into offline planning, where the plan is produced before execution~\cite{saycan}\cite{codeaspolicies2022}\cite{Vemprala2023}\cite{song2023llmplanner}\cite{yang2024vlmtamp}, and online planning, where it is revised during execution~\cite{RePlan}\cite{liu2023reflect}.
SayCan~\cite{saycan} grounds language instructions to actions through reinforcement learning; DELTA~\cite{delta} improves feasibility using scene graphs; others generate BTs by hierarchical or stepwise planning~\cite{llm-bt-1}\cite{llm-bt-2} or integrate classical planners at intermediate stages~\cite{liu2023llmp}.
These methods reflect human intent directly, but require detailed task specifications to be written into the prompt in advance, which is impractical for users without domain expertise.

\begin{figure*}[tb]
    \centering
    \includegraphics[width=1.85\columnwidth]{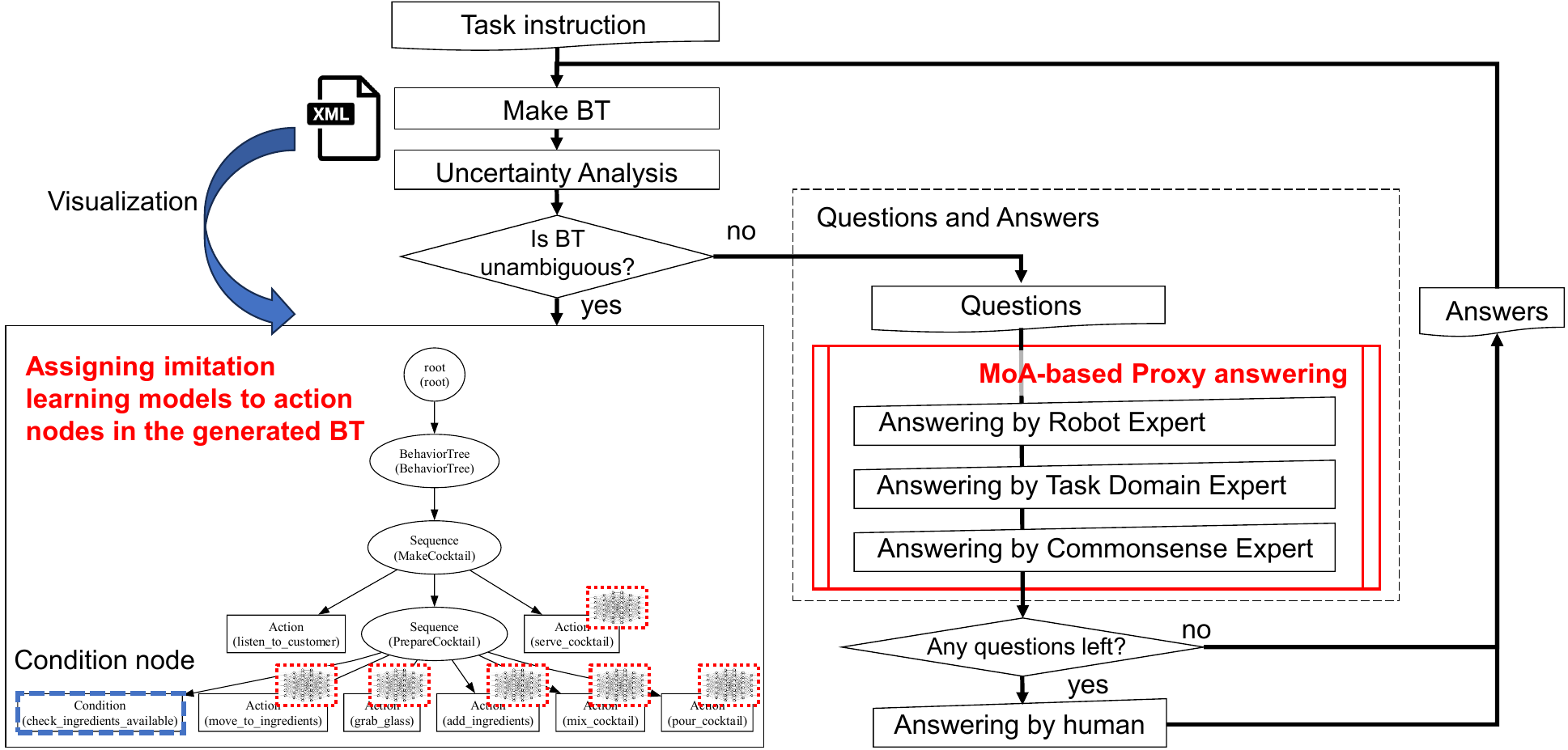}
    \caption{
Overview of the proposed framework.
A task instruction is turned into a BT, whose uncertainty the LLM analyzes to produce clarification questions.
The three expert boxes form the cascade of Sec.~\ref{sec3}: agent $e_k$ holds a prerequisite description $P_k$---the robot's functional specification, the task-domain procedure, and everyday knowledge, respectively---and answers only what $P_k$ entails, forwarding the rest to the next box and finally to the human, the terminal fallback.
Each action node of the refined BT is then bound to a learned policy and the tree is executed on a real robot.
}
    \label{fig2}
\end{figure*}

Interactive task planning alleviates this by having the LLM ask questions and fill in missing information through dialogue~\cite{Huang2022a}\cite{ren}\cite{hori}.
Ren et al. estimate plan uncertainty from instructions and visual observations and selectively generate questions~\cite{ren}; Hori et al. let the LLM analyze uncertainty in its own plan and generate corrective questions for the segments it deems uncertain~\cite{hori}.
The advantage is that only the necessary information is obtained, without detailed task descriptions upfront.

The difficulty is that planners raise many clarification questions, some redundant---asking the ``type of eggs to use'' in a scrambled egg task~\cite{hori}---which increases interaction and operational cost.
Moreover, validation has largely been on short-horizon real-robot tasks~\cite{ren}, so it remains unclear how well such plans support long-horizon execution on physical robots.

Finally, we position our work among architectures that combine multiple LLM agents.
Self-consistency selects the majority answer over sampled reasoning paths~\cite{selfconsistency}; multi-agent debate lets agents critique each other until they converge~\cite{debate}; Mixture-of-Agents organizes proposers into layers and synthesizes their outputs with a dedicated aggregator~\cite{moa}; role-based frameworks such as MetaGPT~\cite{metagpt} and CAMEL~\cite{camel} coordinate role-specific agents through a communication protocol.
All of them expect every agent to answer and then reduce the resulting redundancy to one output, i.e., they target settings where a single best answer exists and must be produced autonomously.
Interactive planning does not have this property: a substantial fraction of clarification questions concern user intent, for which no externally correct answer exists, so answering them autonomously would silently override the user.
We therefore adopt the opposite convention---answer only what the prerequisites entail, abstain otherwise, delegate the residual, and return whatever remains to the human---so that redundancy is avoided rather than resolved (Sec.~\ref{sec3}, Table I).
For BT generation, prior work synthesizes or expands BTs from language instructions~\cite{llm-bt-1}\cite{llm-bt-2}\cite{Cao2023RobotBT}\cite{styrud2024bt} but generally stops at symbolic actions; we instead bind each action node to a concrete imitation learning policy and execute the tree on a physical dual-arm robot, so its retry and fallback structures act on real execution failures.

%% file: sections/3_method.tex
We propose a framework that executes long-horizon tasks on a real robot using a BT generated through interactive task planning, while dynamically switching between multiple motion generation models.
It has two components: BT generation through interactive planning augmented with MoA, which reduces the cost of question--answer interaction; and long-term execution using a motion generation model attached to each BT action node, which tests the feasibility of the generated plan in a real environment.

\subsection{BT Generation via MoA-Based Interactive Task Planning}
Figure~\ref{fig2} shows the overall flow.
The LLM planner first generates a BT from the natural language instruction and analyzes its uncertainty, following~\cite{hori}; this extracts information necessary for execution but absent from the instruction and turns it into clarification questions.
Iterating this dialogue progressively concretizes the specification and yields a BT in XML format that includes conditional branches and retry mechanisms.

\subsubsection{Proxy Response Mechanism via MoA}
Questions are answered not only by the user but also by proxy through MoA, exploiting the commonsense knowledge embedded in LLM agents~\cite{10.5555/3666122.3667509}.
Our agents hold distinct perspectives: robot operational constraints, task specifications, and commonsense reasoning.
Unlike architectures that query every agent and then reduce their redundant outputs to a single response by voting, debate, or a dedicated aggregator~\cite{selfconsistency}\cite{debate}\cite{moa}, our agents are arranged as a cascade in which each agent answers only the questions it can justify from its own prerequisites and delegates the remainder unchanged.

\textbf{Formulation.}
We describe a single dialogue turn; the same procedure repeats at every turn.
Let $Q$ be the set of clarification questions the planner raises, and let $e_1,\dots,e_K$ be the agents in cascade order, agent $e_k$ carrying a prerequisite description $P_k$ that states the knowledge it is entitled to use.
Each agent is specified by two functions: an answerability predicate $\alpha_k$, with $\alpha_k(q)=1$ when $q$ can be answered from $P_k$ alone and $\alpha_k(q)=0$ otherwise, and an answer function $a_k$, applied only where $\alpha_k=1$.
Starting from $R_0=Q$, agent $e_k$ receives the residual $R_{k-1}$ and splits it into the subset $S_k$ it accepts and the subset $R_k$ it forwards unchanged,
\begin{equation}
S_k = \{\,q \in R_{k-1} \mid \alpha_k(q)=1\,\},
\qquad
R_k = R_{k-1}\setminus S_k ,
\end{equation}
returning the answers $A_k=\{(q,a_k(q)) \mid q \in S_k\}$.
The final residual $R_K$ goes to the human user, who supplies $A_H$, so the answer set used to update the plan is $A=A_1\cup\cdots\cup A_K\cup A_H$.
Because $S_k\subseteq R_{k-1}$ and $R_k=R_{k-1}\setminus S_k$, the sets $S_1,\dots,S_K,R_K$ are pairwise disjoint and cover $Q$.
Every question is thus answered exactly once, which is precisely why no voting, weighting, or arbitration stage appears in the framework, and the proxy ratio is simply
\begin{equation}
\rho = 1 - |R_K| / |Q| ,
\end{equation}
the fraction of questions resolved without human input.
Two properties follow.
Coverage is guaranteed: the human is the terminal element, so $R_K$ is always resolved however conservative the agents are.
Cost is at most $K$ calls per turn, and the cascade stops early once $R_k=\emptyset$, whereas layered aggregation costs proposers $\times$ layers $+\,1$ calls~\cite{moa}.
The trade-off is that an early agent's answer is not cross-checked; we return to this in Sec.~\ref{sec5}.
Table I contrasts this design with the architectures of Sec.~\ref{sec2}, and Algorithm~\ref{alg1} gives the procedure.

\begin{table*}[tb]
  \centering
  \label{table:moa_comparison}
  \footnotesize
  \begin{tabular}{l p{4.3cm} l p{3.0cm} l}
    \multicolumn{5}{c}{TABLE I: Design comparison with multi-agent LLM architectures} \\
    \toprule
    Architecture & Aggregation of agent outputs & Abstention & LLM calls per turn & Human role \\
    \midrule
    Self-consistency~\cite{selfconsistency} & Majority vote over samples & No & $N$ samples & None \\
    Multi-agent debate~\cite{debate} & Iterative mutual critique & No & $N \times$ rounds & None \\
    Layered MoA~\cite{moa} & Aggregator LLM synthesizes proposals & No & proposers $\times$ layers $+\,1$ & None \\
    Role-based agents~\cite{metagpt}\cite{camel} & Role-specific message passing & No & Task dependent & None \\
    \textbf{Ours} & \textbf{Not required (disjoint partition)} & \textbf{Yes} & \textbf{$\leq K$ (early exit)} & \textbf{Terminal fallback} \\
    \bottomrule
  \end{tabular}
\end{table*}

\begin{algorithm}[tb]
\caption{MoA-assisted interactive BT generation}
\label{alg1}
\begin{algorithmic}[1]
\REQUIRE instruction $I$, planner $M$, agents $(e_k,P_k)_{k=1}^{K}$
\STATE $B \leftarrow M.\mathrm{GenerateBT}(I)$
\LOOP
\STATE $Q \leftarrow M.\mathrm{AnalyzeUncertainty}(B)$; \ \textbf{if} $Q=\emptyset$ \textbf{then break}
\STATE $R \leftarrow Q$; \ $A \leftarrow \emptyset$
\FOR{$k=1$ \TO $K$ \textbf{while} $R \neq \emptyset$}
\STATE $(A_k, R) \leftarrow e_k(R, P_k)$; \ $A \leftarrow A \cup A_k$ \COMMENT{answer or abstain}
\ENDFOR
\STATE $A \leftarrow A \cup \mathrm{Human}(R)$ \COMMENT{terminal fallback}
\STATE $B \leftarrow M.\mathrm{UpdateBT}(B, A)$
\ENDLOOP
\RETURN $B$
\end{algorithmic}
\end{algorithm}

\begin{figure}[tb]
    \centering
    \includegraphics[width=\columnwidth]{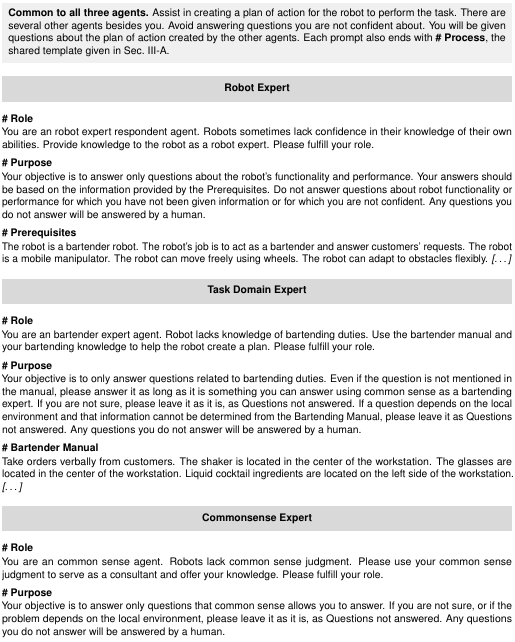}
    \caption{
Prompt design for the three expert agents, with the wording of the deployed prompts left as issued.
The \texttt{\#Prerequisites} block (\texttt{\#Bartender Manual} for the Task Domain Expert) is the prerequisite description $P_k$ that the answerability predicate $\alpha_k$ is evaluated against: each agent answers only what its own $P_k$ entails and forwards the rest.
Sentences common to all three prompts are shown once at the top, and the two long domain blocks are truncated at \textit{[\dots]}.
}
    \label{fig3}
\end{figure}

Every agent receives the same \texttt{\#Process} template as part of its prompt:
\begin{mdframed}
\# Process \\
1. Generate three sections based on the given questions and Prerequisites. \\
a) Analysis of answerability (Analyze and output whether the given question is answerable or not.) \\
b) Answer (Please answer only those questions you are able to answer. Please prefix your answer with a question label.) \\
c) Output of questions not answered (Please output the text you did not answer as is.)
\end{mdframed}
This structure explicitly separates Analysis of answerability, Actual responses, and Identification of unanswered questions.
The three sections realize the quantities above: (a) evaluates $\alpha_k$ over $R_{k-1}$, (b) emits $A_k$ with each answer prefixed by its question label, and (c) reproduces the residual $R_k$ verbatim for the next agent.
Agents are additionally instructed not to answer questions they are unsure of, which biases $\alpha_k$ towards abstention; whatever remains after the last agent goes to the human.

\subsubsection{Example Design of Proxy-Response Agents}
The agents below are one instantiation of MoA; the framework is not tied to a particular configuration. Their prompts are shown in Fig.~\ref{fig3}, all sharing the \texttt{\#Process} section above.

The \textbf{Robot Expert} answers questions on motion capability, executable actions, and sensor configuration, abstaining on anything outside the robot's functional specification.
The \textbf{Task Domain Expert} supplies procedural knowledge specific to the task; we used a bartending agent in Experiment 1 and a smoothie-preparation agent in Experiment 2, the framework itself being domain agnostic.
The \textbf{Commonsense Expert} handles everyday reasoning, abstaining when the answer depends heavily on task context or involves ambiguous judgment.

Proxy responses are not intended to replace human input, but to absorb the questions that are inferable from manuals or general knowledge, or that the user may not know, so that the dialogue retains the information needed to refine the task specification while reducing the number of interactions.

\subsection{Task Execution on a Real Robot Using the Generated BT}
The generated BT separates condition nodes, which describe conditional branching, from action nodes, which describe specific actions, enabling systematic management of state transitions across task stages.

\textbf{BT schema.}
Generated trees follow the BehaviorTree.CPP v4 XML schema (\texttt{BTCPP\_format="4"}). Internal nodes are the control nodes \texttt{Sequence}, \texttt{Selector}, \texttt{Parallel} and the decorators \texttt{Repeat} (\texttt{num\_cycles}), \texttt{RetryUntilSuccessful} (\texttt{num\_attempts}), \texttt{Delay} (\texttt{delay\_msec}); leaves are \texttt{Action} and \texttt{Condition} nodes carrying an \texttt{ID} that names the primitive or predicate and a human-readable \texttt{name}.
The planner is constrained to this vocabulary by its system prompt, so trees load into the execution engine without post-processing.
Figure~\ref{fig_bt_snippet} shows the repetition, verification, retry, and fallback idioms that recur throughout the generated trees.

\begin{figure}[tb]
\centering
\begin{mdframed}
\scriptsize
\begin{verbatim}
<Repeat name="add_strawberries" num_cycles="2">
 <Selector>
  <Sequence>
   <Condition ID="IsStrawberryAvailable"/>
   <RetryUntilSuccessful num_attempts="3">
    <Sequence>
     <Action ID="PutOneStrawberryInBlender"/>
     <Condition ID="IsStrawberryInBlender"/>
    </Sequence>
   </RetryUntilSuccessful>
  </Sequence>
  <Action ID="InformStaffProblem"/>
 </Selector>
</Repeat>
\end{verbatim}
\end{mdframed}
\caption{
Fragment of a generated BT. \texttt{Repeat} realizes the user-specified quantity, the \texttt{Condition} after each \texttt{Action} verifies its outcome, \texttt{RetryUntilSuccessful} re-executes failures, and the \texttt{Selector} falls back to notifying an operator.
}
\label{fig_bt_snippet}
\end{figure}

We assign an individual imitation learning model---Diffusion Policy~\cite{dp} or $\pi_{0.5}$~\cite{pi05}---to each action node, and none to nodes that involve no physical action.
Condition nodes are resolved at run time by a VLM~\cite{Gemini}, which receives the images captured immediately before and after the preceding action node and returns a binary judgement for the predicate named by the node's \texttt{ID} (e.g., \texttt{IsStrawberryInBlender}): SUCCESS if the predicate holds, FAILURE otherwise.
This drives the surrounding control flow: each action is paired with its verifying condition inside \texttt{RetryUntilSuccessful} with \texttt{num\_attempts="3"}, and the enclosing \texttt{Selector} falls back to \texttt{InformStaffProblem} once all attempts fail.

%% file: sections/4_experiments.tex
We conducted two experiments: (1) a quantitative evaluation of the generated BTs, and (2) a long-horizon execution experiment on a real robot using them.

\subsection{Experiment 1}
We first compare BTs generated with and without the proxy-response mechanism in terms of structural and semantic similarity.
The design is internally controlled: the baseline condition is compared against itself to establish the intrinsic spread of the generator, and the proposed condition is then judged against that spread rather than against an external reference.

The task was cocktail preparation, with the LLM given only the abstract instruction ``Make a cocktail.''
Interactive planning had to refine this into the cocktail type (Margarita), the ingredients (tequila, lime juice, orange liqueur), and the procedure (measuring, pouring, mixing, serving).
Since neither procedure nor tools are given at the outset, no executable BT can be built without dialogue.

BTs were generated under two conditions: \textbf{without MoA}, where every question raised during the dialogue was answered by a human, and \textbf{with MoA}, where the cascade answered what it could and the rest was returned to the human.
Generation was performed three times per condition, giving six BTs, using Gemini 2.0 Flash~\cite{Gemini} at temperature 1.0 so that the comparison is made under stochastic generation.

\textbf{Metric 1:}
For structural similarity we use the Tree Edit Distance (TED), the minimum number of node insertions, deletions, and substitutions required to transform one tree into another.
We adopt the definition of Zhang and Shasha~\cite{K.Zhang1989} with the normalization of Li and Zhang~\cite{Li2011}, which handles cases where the triangle inequality does not hold:
\begin{equation}
d_{\mathrm{N\text{-}TED}}(T_1, T_2)
=
\frac{2 \cdot \mathrm{TED}(T_1, T_2)}
{\alpha (|T_1| + |T_2|) + \mathrm{TED}(T_1, T_2)}
\end{equation}
where $|T|$ is the number of nodes and $\alpha$ the maximum edit cost, here 1 for all operations. The measure lies in $[0,1]$, smaller being more similar.

\textbf{Metric 2:}
For semantic similarity, each node is expressed in natural language and embedded with Sentence-BERT~\cite{Reimers2019}. For every node of one BT we take the maximum cosine similarity against all nodes of the other, and average over nodes; the measure lies in $[0,1]$, larger being more similar, and is asymmetric in its two arguments.

\subsection{Experiment 2}
To evaluate whether generated BTs function as long-horizon executable plans in real environments, we conducted a smoothie-making task on the tabletop dual-arm robot ALOHA~\cite{act}, which must reflect user preferences obtained through dialogue such as the number of fruits.
The task consists of (i) inserting multiple types of fruit (strawberry, banana, kiwi) into the blender, (ii) closing the blender lid, (iii) turning on the blender switch, (iv) waiting, and (v) turning off the blender switch.
The fruits are food replicas, and the blender is left unpowered during autonomous operation for safety, so the ingredients are not physically blended.
See the supplemental video for the executed behavior.

For actions (i), (ii), (iii), and (v), we trained both $\pi_{0.5}$~\cite{pi05} and Diffusion Policy~\cite{dp}, and assigned the model with the highest success rate to the corresponding BT node.
For demonstration data collection, we gathered 400 sequences for strawberry and 300 sequences each for banana and kiwi in subtask (i), and 200 sequences each for subtasks (ii), (iii), and (v).
Demonstrations were recorded by teleoperation at 30\,fps with three RealSense cameras (one overhead and one per wrist, $640\times480$), and the action space is the 14-dimensional joint configuration of the two 7-DoF arms.
Diffusion Policy was trained separately for each subtask, whereas a single $\pi_{0.5}$ model was trained jointly on the three fruit datasets and conditioned at inference time on the language instruction of the corresponding action node.
Batch size was 64 for $\pi_{0.5}$ and 16 for Diffusion Policy.
Diffusion Policy was trained for 200,000 iterations; for $\pi_{0.5}$ we trained up to 400,000 and selected the checkpoint per action node, giving 400,000 iterations for strawberry and 200,000 for banana and kiwi.
At inference time, $\pi_{0.5}$ was run in \texttt{bfloat16} with 10 denoising steps, and each action node was given an execution budget of 20\,s.
All other training and inference parameters followed the default settings provided in LeRobot library~\cite{lerobot2024}.
Interactive planning in this experiment used Gemini 2.5 Flash~\cite{Gemini} at temperature 0, chosen to prioritize the stability of the generated plans.

%% file: sections/5_results.tex
\subsection{Evaluation of Generated BTs}
\label{sec5_bt}
From the recorded dialogue logs, the system generated an average of 37 clarification questions per BT. 
Among these, the MoA-based proxy response mechanism answered an average of 10 questions, corresponding to a proxy ratio of $\rho \approx 0.27$ of all generated questions.
The rest were returned to the human by the terminal fallback, so no question was left unresolved.

\begin{table}[tbp]
  \centering
  \label{table:bt_ted}
  \begin{tabular}{cc|ccc|c}
    \multicolumn{6}{c}{TABLE II: Normalized TED against BTs w/o MoA ($\downarrow$)} \\
    \hline
    & & A & B & C & Ave. \\
    \hline
    \multirow{3}{*}{w/o MoA} & A & - & 0.807 & 0.817 &  \\
     & B & 0.807 & - & 0.776 & 0.800 \\
     & C & 0.817 & 0.776 & - &  \\
    \hline
    \multirow{3}{*}{w/ MoA} & A' & 0.813 & 0.774 & 0.808 & \\
     & B' & 0.828 & 0.750 & 0.762 & 0.812 \\
     & C' & 0.812 & 0.875 & 0.887 & \\
    \hline
  \end{tabular}
\end{table}

\begin{table}[tbp]
  \centering
  \label{table:bt_similarity}
  \begin{tabular}{cc|ccc|c}
    \multicolumn{6}{c}{TABLE III: Node similarity against BTs w/o MoA ($\uparrow$)} \\
    \hline
    & & A & B & C & Ave. \\
    \hline
    \multirow{3}{*}{w/o MoA} & A & - & 0.640 & 0.595 &  \\
     & B & 0.736 & - & 0.618 & 0.664 \\
     & C & 0.731 & 0.666 & - &  \\
    \hline
    \multirow{3}{*}{w/ MoA} & A' & 0.757 & 0.726 & 0.664 & \\
     & B' & 0.724 & 0.677 & 0.674 & 0.697 \\
     & C' & 0.701 & 0.657 & 0.692 & \\
    \hline
  \end{tabular}
\end{table}

\begin{figure*}[tb]
    \centering
    \includegraphics[width=1.9\columnwidth]{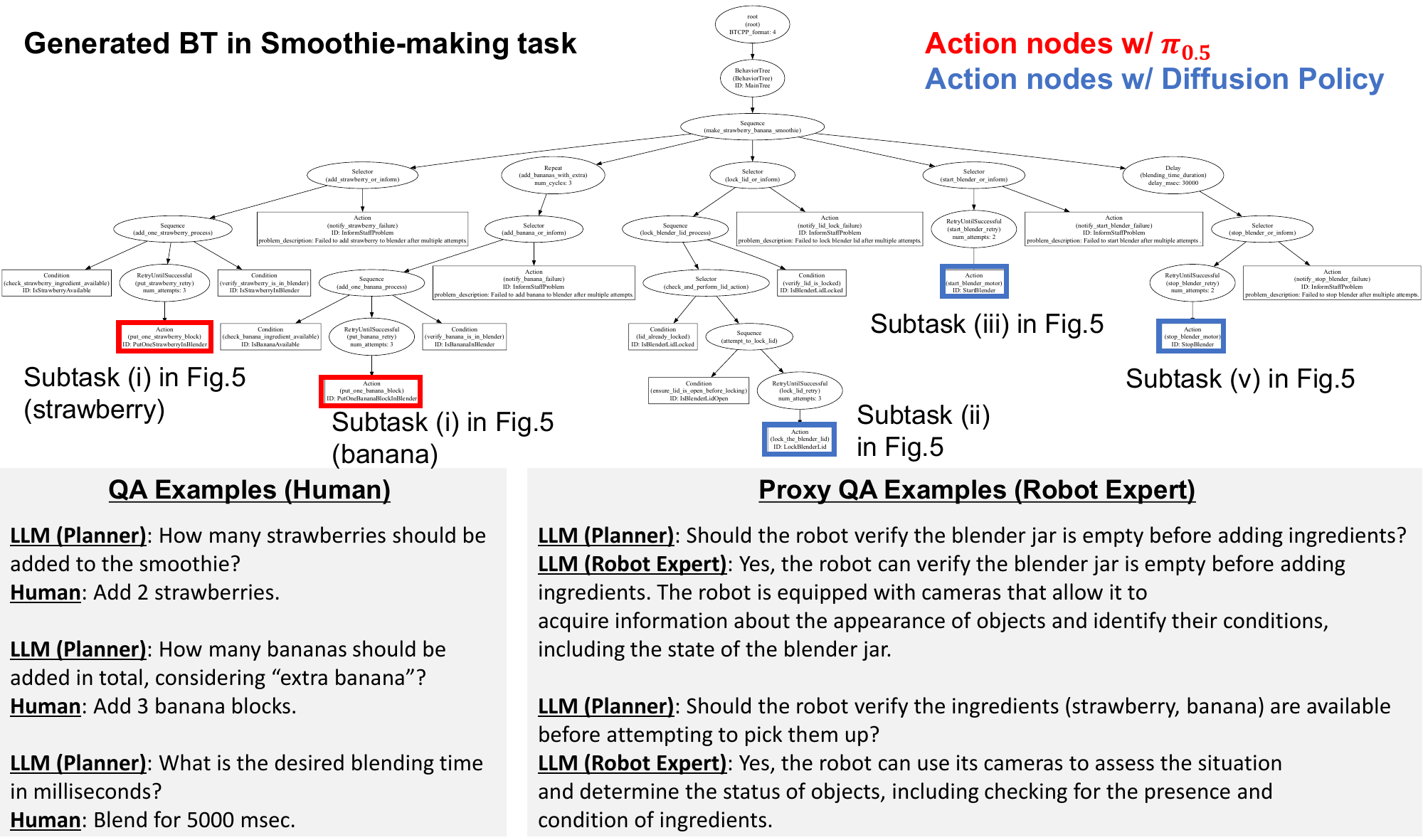}
    \caption{
BT generated through MoA-assisted interactive planning for the smoothie-making task, with example question--answer interactions. Robot-related questions are answered by the Robot Expert, user-preference questions by the human, and action nodes are assigned either Diffusion Policy or $\pi_{0.5}$.
}
    \label{fig5}
\end{figure*}

Table II reports the normalized TED among baseline BTs (trials A--C) and between baseline and proposed BTs (trials A'--C').
The key comparison is not between two independent methods but between a condition and the generator's own spread: repeated baseline generations already differ from one another by $0.800 \pm 0.021$ over the three distinct baseline pairs, and the proposed BTs differ from the baseline by $0.812 \pm 0.047$ over the nine cross pairs.
The gap between the two means, $0.012$, is smaller than the standard deviation of either distribution, so proxy answering moves the generated tree no further than re-running the baseline generator does.

Table III shows the same pattern for node embedding similarity: 0.664 among baseline BTs against 0.697 between baseline and proposed BTs, the proposed condition being marginally the more similar of the two.
Both metrics place the proposed BTs within the baseline's own generation variance: proxy answering removes human effort without displacing the resulting plan.

\begin{figure*}[tb]
    \centering
    \includegraphics[width=1.8\columnwidth]{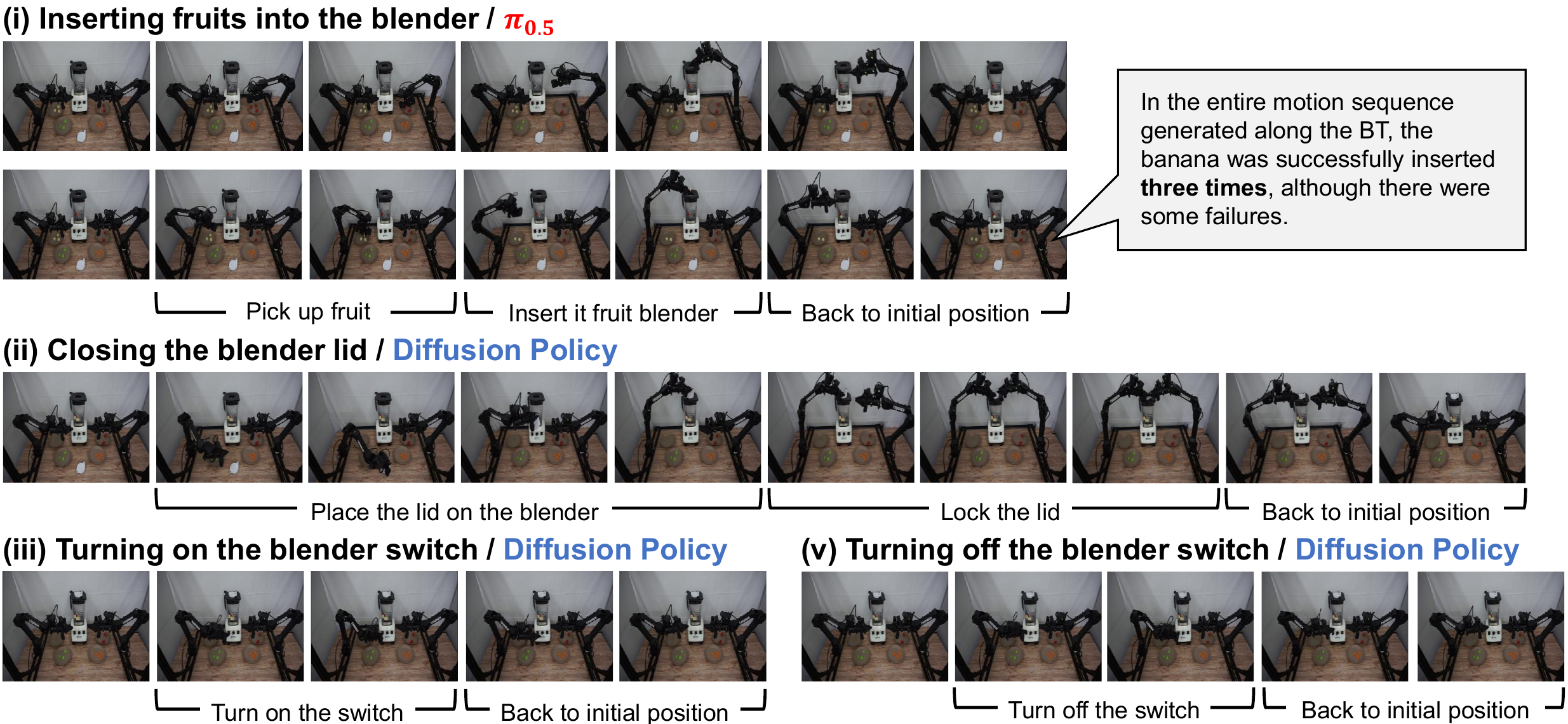}
    \caption{
Real-robot execution sequence guided by the generated BT, switching between $\pi_{0.5}$ and Diffusion Policy according to the action node. 
}
    \label{fig6}
\end{figure*}

\subsection{Long-Horizon Task Execution on a Real Robot}
Figure~\ref{fig5} shows the BT generated for the smoothie-making task with its question--answer exchanges; a complete tree was obtained at the second planning iteration. 
During BT generation, a total of five clarification questions were produced.
Among these, two were answered by the Robot Expert agent, none by the Task Domain Expert or the Commonsense Expert, and the remaining three were returned to the human user ($\rho = 0.4$).

\begin{table}[tb]
\centering
\label{tab:action_success_rate}
\footnotesize
\setlength{\tabcolsep}{4pt}
\begin{tabular}{l c c c}
\multicolumn{4}{c}{TABLE IV: Success rates of actions (10 trials)} \\
\toprule
 & (a) DP$^{*}$ & (b) $\pi_{0.5}$ & (c) $\pi_{0.5}$ \\
Action & & subtask (i) & all subtasks \\
\midrule
Insert strawberry & 0/10 & 3/10 & 0/10 \\
Insert banana & 0/10 & 6/10 & 6/10 \\
Insert kiwi & 0/10 & 6/10 & 2/10 \\
Close blender lid & 7/10 & --- & 0/10 \\
Switch blender on & 10/10 & --- & 1/10 \\
Switch blender off & 10/10 & --- & 2/10 \\
\bottomrule
\end{tabular}
\begin{flushleft}
$^{*}$ Diffusion Policy, trained independently for each subtask.
\end{flushleft}
\end{table}

Table IV reports success rates over 10 trials per action.
The narrowly scoped models were best for every action: Diffusion Policy trained per subtask for the blender operations (IV-a), and $\pi_{0.5}$ restricted to subtask (i) for fruit insertion (IV-b), which likely benefits from tabletop fruit manipulation in its pretraining data.
The single $\pi_{0.5}$ trained on all actions degraded sharply on several of them, indicating that multi-task learning here needs substantially more demonstrations.
We therefore used $\pi_{0.5}$ trained on action (i) for fruit insertion and Diffusion Policy for actions (ii), (iii), and (v).

Figure~\ref{fig6} shows the executed sequence.
The system switched between motion generation models and VLA prompts according to the required action and completed the task from start to finish; occasional fruit-grasping failures were re-executed by the retry structure, and the user-specified repetition counts were realized by the \texttt{Repeat} decorators, without interrupting the task.
Retry raised effective completion above the per-attempt rates of Table IV, but the intrinsic accuracy of each subtask remains the limiting factor.

\subsection{Failure Modes and Applicability Boundaries}
\label{sec5_failure}
Table V lists the modes, their boundaries, and our mitigations. Two properties shape that list: with one policy per action node, a failure is attributable to a named node, and with the human terminating the cascade and \texttt{InformStaffProblem} the tree, every mode has a defined behavior.

\textbf{Manipulation.} Two modes arise at the action nodes. The dominant is grasping objects whose pose and shape vary: strawberry insertion reached only 3/10 and Diffusion Policy failed on all three fruits, yet the same models were reliable on the fixed-geometry blender operations. The boundary is object variability, not model family---the more tractable diagnosis, implicating the demonstration set rather than the policy architecture. The second mode appears when one policy covers heterogeneous subtasks: the jointly trained $\pi_{0.5}$ collapsed on those with the fewest demonstrations, indicating interference rather than lack of capacity---which per-node binding let us avoid in deployment.

\textbf{Horizon amplification.} Assuming independent attempts, a node with per-attempt success $p$ under three \texttt{RetryUntilSuccessful} attempts succeeds with probability $1-(1-p)^3$. Applying this to the rates of Table IV over the tree of Fig.~\ref{fig5} gives an estimated end-to-end success of $\approx 0.34$, dominated by the two strawberry insertions ($p=0.30$); since each repetition enters as an exponent, the tree drops to $\approx 0.06$ if six strawberries are asked for. Deepening retry does not close the gap (six attempts lift that node only to $0.88$), but because the tree is explicit the product can be evaluated before execution, so a node below threshold can be escalated through the \texttt{InformStaffProblem} fallback rather than failing silently.

\textbf{Dialogue.} With one agent per question, an early error propagates unchallenged, so each $P_k$ is critical: one that has drifted from the robot yields confident wrong answers. The profile is nonetheless asymmetric in the favorable direction: over-conservative abstention merely degrades the cascade towards the all-human baseline at a lower $\rho$, whereas the damaging mode is an agent answering a preference question and overriding the user silently. We did not observe it in Experiment 2: the three questions returned to the human were exactly the preference ones, which we credit to the abstention-first instruction rather than any guarantee. Two cheap safeguards follow: grounding each $P_k$ in a machine-readable platform description, and confirming the proxy answers in one batch (Sec.~\ref{sec5_bt}).

\textbf{Verification.} A false FAILURE costs at most the remaining retry budget before the fallback, whereas a false SUCCESS advances the tree on an unsatisfied precondition and is bounded by nothing---hence returning FAILURE when uncertain.

\begin{table*}[tb]
  \centering
  \label{table:failure_modes}
  \scriptsize
  \setlength{\tabcolsep}{4pt}
  \begin{tabular}{p{2.5cm} p{4.3cm} p{3.7cm} p{4.9cm}}
    \multicolumn{4}{c}{TABLE V: Failure modes, applicability boundaries, and mitigation directions} \\
    \toprule
    Failure mode & Evidence & Boundary & Mitigation direction \\
    \midrule
    Grasp failure on variable objects & Strawberry 3/10 with best policy; DP 0/10 on all fruits & High pose and shape variability & Allocate demonstrations by object variance \\
    Cross-task interference & Jointly trained $\pi_{0.5}$: lid 0/10, switches 1--2/10 & One policy over heterogeneous subtasks & Widen scope only with a matching demo budget \\
    Horizon amplification & Estimated end-to-end success $\approx 0.34$ & Long sequences with a weak node & Estimate at planning time; escalate weak nodes \\
    Unchecked proxy answer & One agent per question; no second opinion & Stale or incorrect prerequisite $P_k$ & Ground $P_k$ in a machine-readable platform spec \\
    Over-answering preference questions & Not observed in Exp.~2 & Factual and preference questions overlap & Batch confirmation before BT generation \\
    Condition-node misjudgement & Not evaluated separately & Visually ambiguous or occluded predicates & Return FAILURE when uncertain: the bounded error \\
    \bottomrule
  \end{tabular}
\end{table*}

\subsection{Limitations and Generalization}
\label{sec5_limitations}
The cascade uses a fixed, hand-authored agent order. Fusion is inapplicable rather than missing---the partition leaves nothing to reconcile---but the fixed order is a genuine restriction: cost grows linearly in $K$, so a large expert pool would need a routing step---which changes which $\alpha_k$ are evaluated, not the disjointness of the partition; a second opinion could be restored on the same terms, for factual questions only. Planning is also offline: the BT is fixed before execution, so unanticipated situations are handled only by structures already in the tree---though the tree is where this could be relaxed, by regenerating a subtree at the failing node.

The components generalize unevenly. The BT schema, Algorithm~\ref{alg1}, and the \texttt{\#Process} template are task independent and were reused unchanged in both experiments; what must be re-authored per domain is each $P_k$, the action-node to policy binding, and the demonstration data. Both experiments are drink preparation, so this is reuse within a domain rather than across domains. The demonstration data dominates the cost and bounds the achievable horizon: a new task is cheap on the planning side and expensive on the manipulation side. That cost was lowest where $\pi_{0.5}$ could draw on related tabletop manipulation in its pretraining data (Table IV), so broader pretraining is the likelier route into new domains.

%% file: sections/6_conclusion.tex
This study presented a framework for long-horizon robot task execution that combines interactive LLM-based planning, MoA answering, and BT generation.
We formulated proxy answering as an abstention-based delegation cascade: the agents partition the question set disjointly and the human is the terminal fallback, so answer fusion is unnecessary while every question is still covered. Quantitative evaluation showed that the generated BTs remain within the baseline generator's own spread under both normalized Tree Edit Distance and embedding-based similarity, so proxy answering removes human effort without displacing the resulting plan.
Real-robot experiments showed the generated BTs execute directly in a physical environment: by binding a different imitation learning model to each action node and using the retry structures of the tree, the robot completed a multi-step smoothie-making task despite occasional execution failures.
The failure analysis showed that the applicability boundary is set by manipulation reliability rather than by planning: retry raises per-node success, but the weakest node dominates the achievable horizon.
Future work includes scaling the MoA framework to broader task domains, incorporating automatic agent selection so that the cost of the cascade does not grow with the size of the expert pool, and improving the per-node policies that currently bound end-to-end performance.